\documentclass[conference]{IEEEtran}
\IEEEoverridecommandlockouts
\usepackage{cite}
\usepackage{amsmath,amssymb,amsfonts}
\usepackage{algorithmic}
\usepackage{graphicx}
\usepackage{textcomp}
\usepackage{xcolor}
\usepackage{tikz}
\usepackage{textcomp}
\usepackage{hyperref}
\usepackage{lipsum}
\def\BibTeX{{\rm B\kern-.05em{\sc i\kern-.025em b}\kern-.08em
    T\kern-.1667em\lower.7ex\hbox{E}\kern-.125emX}}
\begin{document}

\title{Deep and Probabilistic Solar Irradiance Forecast at the Arctic Circle\\

}

\newcommand\copyrighttext{%
  \footnotesize \textcopyright 2024 IEEE. Personal use of this material is permitted.
  Permission from IEEE must be obtained for all other uses, in any current or future 
  media, including reprinting/republishing this material for advertising or promotional 
  purposes, creating new collective works, for resale or redistribution to servers or 
  lists, or reuse of any copyrighted component of this work in other works. 
  }%DOI: \href{<http://tex.stackexchange.com>}{<DOI No.>}}
\newcommand\copyrightnotice{%
\begin{tikzpicture}[remember picture,overlay]
\node[anchor=south,yshift=10pt] at (current page.south) {\fbox{\parbox{\dimexpr\textwidth-\fboxsep-\fboxrule\relax}{\copyrighttext}}};
\end{tikzpicture}%
}

\author{\IEEEauthorblockN{Niklas Erdmann\textsuperscript{1}, Lars Ø. Bentsen\textsuperscript{1}, Roy Stenbro\textsuperscript{2}, Heine N. Riise\textsuperscript{2}, Narada Warakagoda\textsuperscript{1,3},\\ and Paal Engelstad\textsuperscript{1} }\\

\IEEEauthorblockN{\textsuperscript{1}Department of Technology Systems, University of Oslo, 2007 Kjeller, Norway \\
\textsuperscript{2}Institute for Energy Technology (IFE), 2007 Kjeller, Norway\\
\textsuperscript{3}Norwegian Defense Research Establishment (FFI), 2007 Kjeller, Norway}}
% \and
% \and

% }

\maketitle
\copyrightnotice
%abstract to 300 words 
\begin{abstract}
%Solar energy production can be dynamic and unreliable due to changing weather conditions. 
% The main driver of revenue based on solar energy production in Norway is the Day-ahead energy market. Energy production forecast inaccuracies will incur a loss during the settlement on the market, thus higher forecasting accuracy may cause higher return and perhaps higher reinvestment into the technology. While solar energy production data is not readily available in Norway, solar irradiance observations are, which can be seen as directly related to energy production. Weather forecasting, like for 
%Solar irradiance forecast, has been a difficult task and traditionally dominated by numerical models. In recent times however, this shifted to successful applications of Deep learning. One weakness of such AI based approaches is the missing trustworthiness of results. 
Solar irradiance forecasts can be dynamic and unreliable due to changing weather conditions.
Near the Arctic circle, this also translates into a distinct set of further challenges.
This work is forecasting solar irradiance with Norwegian data using variations of Long-Short-Term
Memory units (LSTMs). In order to gain more trustworthiness of results,
the probabilistic approaches Quantile Regression (QR) and Maximum Likelihood (MLE)
are optimized on top of the LSTMs, providing measures of uncertainty for the results.
MLE is further extended by using a Johnson's SU distribution, a Johnson's SB distribution, and a Weibull distribution in addition to a normal Gaussian
to model parameters. Contrary to a Gaussian, Weibull, Johnson's SU and Johnson's SB can return skewed distributions,
enabling it to fit the non-normal solar irradiance distribution more optimally. The LSTMs
are compared against each other, a simple Multi-layer Perceptron (MLP), and a smart-persistence estimator. The proposed LSTMs are found to be more accurate than smart persistence and the MLP for a multi-horizon, day-ahead 
(36 hours) forecast. 
The deterministic LSTM showed better root mean squared error (RMSE), but worse mean absolute error (MAE) than a MLE with Johnson's SB distribution. Probabilistic uncertainty estimation is shown to fit relatively well across the distribution of observed irradiance. While QR shows better uncertainty estimation calibration, MLE with Johnson's SB, Johnson's SU, or Gaussian show better performance in the other metrics employed. Optimizing and comparing the models against each other reveals a seemingly inherent trade-off between point-prediction and uncertainty estimation calibration.  
\end{abstract}

% \begin{IEEEkeywords}
% Forecasting, Irradiance forecast, Photovoltaic, Deep learning, Probabilistic forecasting, Multihorizon, Norway
% \end{IEEEkeywords}

\section{Introduction and Background}\label{sec:intro}

Accurate solar irradiance forecasting is an area of continuous interest to advance the capability of forecasting photovoltaic (PV) energy production. Solar irradiance can be used to model PV energy in a straightforward manner~\cite{widenVariabilityAssessmentForecasting2015}. In addition, it is much simpler to gain data on solar irradiance, often associated with public weather data, compared to PV power production, often associated to private companies. 
The day-ahead energy market is an important tool for related producers, however, settlement of this market results in losses for producers depending on their error in energy production forecast~\cite{klyveValueForecastsPV2023a}. A higher accuracy may lead to higher profits for producers and provide incentive to reinvest in more PV energy production.

Forecasts of solar irradiance using machine learning has been investigated at length in recent years, as machine learning techniques has been shown to be able to beat traditional numerical weather prediction models (NWPs) in a short-horizon forecasts of hours into the future~\cite{kumariDeepLearningModels2021}.
Machine learning techniques are often tested on short-term or ultra short-term forecasts (i.e. subhourly, and hourly up to six hours). 
 Performance comparisons are getting more blurry between machine learning approaches (but also between NWP and ML) in a less short-term setting, i.e. forecasting day(s) ahead~\cite{krishnanHowSolarRadiation2023}.
In addition to this, a higher forecast horizon is more beneficial for the day-ahead market, with the potential to lead to more profitable solar energy production for investors~\cite{klyveValueForecastsPV2023a}. This paper will use a forecast horizon of 36 hours, as there seem to be a need for more investigation in longer horizons in combination with machine learning techniques. 

Machine learning architecture wise, Long-Short Term Memory units (LSTMs) have been continually shown to perform well  in various forecasts, including solar irradiance (e.g.~\cite{srivastavaComparativeStudyLSTM2018,jailaniInvestigatingPowerLSTMBased2023}) presenting them as a natural choice for this work.
%TODO Can add more now

A major shortcoming of point predictors is the inability to tell how likely a given result is correct. Modelling uncertainty has become an increasingly important topic in recent years, not only in solar irradiance forecasting. Showing the uncertainty of a result is linked to higher trust of users into that result, facilitating more practical use of the related technology. Measures of uncertainty can be obtained by a family of probabilistic modeling approaches, some of which have been employed in in solar irradiance forecasts already.
This work focuses on probabilistic outputs and loss functions, such as Quantile regression neural networks (QRNNs) and Maximum Likelihood Estimation (MLE) which can be used in conjunction with neural networks to model the uncertainty of solar irradiance forecasts. 
The benefit of such approaches is that they are a relatively simple extension to a neural network, but still integrated to the architecture. Output is still harvested directly from a trained network, which instead of making a point prediction, estimates a probability distribution based on its internal representation of the data.

Quantile regression (QR) is a nonparametric approach to directly model a set of quantiles of a probability distribution. It has been found effective in solar irradiance forecast with a linear estimator~\cite{lauretProbabilisticSolarForecasting2017} and neural network based estimators~\cite{davidComparisonIntradayProbabilistic2018}. More recently, QR has been used with a combination (hybrid) of a Convoluted Neural Network (CNN) and Long Short-Term Memory Unit (LSTM) for solar irradiance forecasts~\cite{sansineHybridDeepLearning2023}.
% MLE:
MLE is used to fit the parameter of a distribution function to the underlying distribution of the data. The parametrized distribution function used for this process can be chosen freely and may benefit from making an informed decision based on which functions may be able to best fit the underlying true distribution of the data.
This represents a naturally included prior in estimating the true distribution which may benefit the end result of the forecast~\cite{awasthiBenefitsMaximumLikelihood2021}.
MLE has been used in solar irradiance forecasting mainly to compute the distribution of residuals~\cite{heProbabilisticSolarIrradiance2020, juniorUseMaximumLikelihood2015a}. The authors did not find any work using MLE in conjunction with neural networks for solar irradiance forecasts.
MLE thrives or breaks from the choice of modelling distribution. As such, there exists a set of reoccurring MLE target probability distributions in solar irradiance forecasts. 
The standard Gaussian distribution is a natural choice in MLE approaches, however in wind and even solar forecast, previous work examined more distributions. Specifically interesting are distribution that are able to model skewness, such as the Weibull distribution~\cite{kamComparativeWeibullDistribution2021} or the Laplace distribution~\cite{juniorUseMaximumLikelihood2015a}.  % beta distribution mention???
An potential downside of this set of distributions is that they may only model one direction of skewness and are also bounded. The family of Johnson's transformations can be used to create distributions which may model skew in any direction and can either be bounded or unbounded~\cite{johnsonog}. The unbounded version, Johnson's SU distribution, has been shown to work well for wind forecasts~\cite{bentsenRelativeEvaluationProbabilistic2024}, but has not been investigated in the context of solar irradiance forecasts. Johnson's SB distribution is bounded between a minimum and maximum value and has not been used in solar irradiance or wind forecasts to our knowledge.
In order to test the performance or different distributions in solar irradiance forecasts, this work is using the Gaussian, Weibull, Johnson's SU, and Johnson's SB distributions. Weibull can model skew in one direction, but is also bounded in one direction. The two Johnson's distribution can model skew (in both directions) and kurtosis, while either being completely unbounded, or bonded in both directions. The Gaussian however is also unbounded, but can not model skewness.

Lastly, most work regarding solar irradiance forecasts has been conducted using data from temperate regions, or regions close to the equator. 
Forecasting solar irradiance close to the Arctic circle (in this case with Norwegian data) brings its own set of problems compared to other locations on the globe: A drastically changing day-and night circle not only leads to less sunlight during the days in winters, but also shorter days in general, leading to relatively less available data compared to the summer months. Additionally, geostationary satellites lack coverage into the Arctic circle, limiting solar irradiance data availability. Low solar elevation and the prevalence of snow may lead to further issues, such as satellite misinterpretation of snow as clouds, or not accurately quantifying ground albedo. Both issues contribute to errors related to satellite based estimation of solar irradiance. Here, clear sky irradiance is injected into model output, enabling models to only needing to learn the difference in clear sky and actual irradiance. This is thought to mitigate variance of sunlight due to the changing day-night cycle, and the lack of data in the winter months. 
% conclusion on this?????
\subsection{Contributions}\label{sec:contributions}
This work strives to bring tried and trusted machine learning techniques in line with the challenges of solar irradiance forecasts close to the Arctic circle while building upon more practically oriented design choices: 
\begin{itemize}
    \item A  long-short term memory unit (LSTM) is optimized to  forecast solar irradiance in a deterministic way up to 36 hours into the future.
    \item A 36 multi-horizon is not as often investigated as one hour-ahead forecasts, but vastly more useful for application in the day-ahead market.
    \item Forecasting solar irradiance from the Arctic cycle with clear sky injection into model output. 
    \item The optimized LSTM is extended with three probabilistic approaches: Quantile regression (QR),  Maximum Likelihood estimation (MLE) fitting to a Gaussian distribution, Johnson's SU, and SB distribution, and a Weibull distribution.
    \end{itemize} 

\section{Method}\label{sec:method}
\subsection{Data}\label{sec:data} % data location and preprocessing
Hourly observations of surface downwelling shortwave radiation from 20 stations around the southern area of Norway across 5 years (2016-2020) are designated as input to the models in combination with irradiance data from CAMS-RAD (an European satellite-image based model), estimated weather features from NORA3 (a hind-cast numerical weather model, spanning the sea and land area around Norway), and a clear sky index from CAMS McClear. Of these inputs, values corresponding to the coordinates of the (first) 20 observation stations are extracted. 
The weather data extracted from NORA3 are an integral of surface net downward shortwave flux, temperature, relative humidity, cloud area fraction, surface air pressure, wind direction and speed, snowfall, and precipitation amount. 
In addition, time variables for hour-of-day, day-of-week and week-of-year are embedded and added to the input. This embedding was done by decomposing the input into a sine and cosine component as shown in~\ref{eq:time_embedding} with $t$ being the input time and $d$ being the embedded time.
\begin{equation}\label{eq:time_embedding}
    \begin{aligned}
    d_{\text{sin}} = \sin\left(\frac{2\pi t}{24}\right),
    d_{\text{cos}} = \cos\left(\frac{2\pi t}{24}\right)
    \end{aligned}
\end{equation}
With the addition of embedded time, the input dimensionality summation results in $20(1(Observation) + 9(NORA3) + 1(CSI) + 1(Cams)) + 6(embedded\_time) = 246$ input features. 
The input is set to forecast solar irradiance values in a horizon of 1-36 hours from station eleven (due to it being centrally located and featuring the least amount of missing data).

\subsection{Preprocessing}\label{sec:preprocessing}
As part of the data preparation, the missing values are replaced with zeros, as the target variable only features a negligible amount, not warranting data interpolation. The data is then normalized. Normalization has been tested with a min-max and standard scaler. Performance was found to be similar between both, but the min-max scaler is eventually chosen for its ability to avoid negative values in the output. This is relevant for MLE using the Weibull distribution, as it is not defined for negative parameter values. Batch-normalization was also tested, but found to not improve performance. 

Due to the relatively high amount of input features, and some of the features (in particular some weather features) only being weakly linearly correlated to irradiance observations, performance testing using a dimensionality-reduced form of the data set are also conducted (using a PCA and an Auto-encoder). They are not included here, as they seemed to generally perform worse than without reduction.

As a last step, the data is split into a train, validation (for hyper-parameter optimization) and test (for final optimized model test) set. The split is done in years to be able to test on a whole year of data. Correlations between years are examined and the least correlated years (one and two) used for testing and validation (respectively), to increase (instead of decrease) forecast difficulty. Years three, four and five are used for training. 

%TODO:  cv minus,
\subsection{Deterministic LSTM} %TODO check Lars papers for how to do models and metrics best
% Talk about architecture design here. Window, timesteps, linear transform output, clear sky injection, persistence connection, learnable parameter scaling
LSTMs, first introduced by~\cite{hochreiter1997long} are an extension on traditional Recurrent Neural Networks (RNNs). LSTMs can remember information over longer periods, but still not suffer from vanishing/exploding gradient problems. The LSTM architecture is very prevalent in the irradiance forecast literature, thus the basic setup will be omitted here for brevity. See for example~\cite{srivastavaComparativeStudyLSTM2018} for a detailed description of LSTMs in solar irradiance forecasting.

In this work the LSTM input is a window of a number of past data points. The LSTM is slid over the window and the final return states are fed into a linear layer, mapping to the output horizon of 36 hours. In other words, input has the shape $x \in \mathbb{R}^{B\times W \times D} $ with $B$ being batch size, $W$ being window size and $D$ being the number of features. The last hidden activation states are then fed into a linear transformation layer with output $\hat{y} \in \mathbb{R}^{B \times P}$ in which $P$ is the size of the output horizon.
% \begin{equation}
%     H_{t} = LSTM(x_{t})
% \end{equation}
% \begin{equation}
%     \hat{y} = Linear(h_{t})
% \end{equation}
In extension to this basic setup, a number of further modifications are implemented and tested for performance: Future clear sky values are arguably always available in real world settings thus thought to be injected directly in the output of the model, providing them the ability to only needing to differentiate between the theoretical maximum of solar irradiance and the actual observations. This is done by combing min-max scaled clear sky $cs$ with (a part of the) output $\hat{y}$ and a learnable parameter $\alpha$ as shown in Equation~\ref{eq:clear_sky_inject}.
\begin{equation}\label{eq:clear_sky_inject}
    \hat{y}' = \hat{y} + \alpha * cs
\end{equation} 
This is thought to be very beneficial in helping the model bridge the big magnitude and day-length differences between summer and winter close to the arctic circle. $\hat{y}$ is set to different parts of the output depending on the model employed. For the deterministic LSTM, $\hat{y}$ is set to the final point output. In the probabilistic models, $\hat{y}$ is set the mean for Gaussian and the 0.5 quantile for QR. In the case of Weibull, Johnson's SU and Johnson's SB, injection of $cs$ required introduction to the parameters, thus is explained in their respective sections in Section~\ref{sec:distributions}.

The LSTM is also extended with a learnable-parameter-scaled persistence connection, adding the previous 24 hours of irradiance observation to the output (with the first 12 hours being repeated for hour 24-36). Instead of the clear sky injection, it was also tested to subtract clear sky values from the input and expected output irradiance, to enable the LSTM not having to learn only the difference between clear sky and actual irradiance.
 In preliminary testing, only the injected clear sky values improved performance, thus the persistence connections and subtracting of clear sky was disabled subsequently.

\subsection{Probabilistic LSTMs}\label{sec:probabilistic}
In order to include an estimation of uncertainty for a forecast, the deterministic LSTM is extended with a parametric and non-parametric probabilistic approach. Instead of a point-value forecast of the future irradiance data, probabilistic methods will generate a probability distribution for each forecasted time-point. Thus, with the considered methods, the main change are the output shape and the loss function, as the LSTM needs to be adapted to output a probability distribution instead of a point value.

\subsubsection{Quantile Regression}\label{sec:QR}
QR models are trained to predict the $q$th quantiles of the target distribution, with $q \in Q$ which can be seen as the output of the QR model. The values of $q$ represent the probability that the true value is smaller than the quantile $q$. The output of QR then is a set of quantiles, $Q$, e.g. $Q = \{0.05, 0.25, 0.5, 0.75, 0.95\}$.

Thus, the size of the output is extended such that $ y \in \mathbb{R}^{B \times P \times |Q|} $. The typical loss function for QR is known as Pinball loss, or quantile loss, defined as:
%\begin{equation}
\begin{align} %TODO needs adaption to rest later
    L_{QR}(y, \hat{y}) &= \frac{1}{BW}\sum_{b \in B} \sum_{w \in W} \sum_{q\in Q} \rho_q(u)(y_{b,w} - \hat{y}_{b,w,q})  \label{eq:pinball1} \\ 
    \rho_q(u) &= \begin{cases} q u & \text{if } u \geq 0\\ (1-q)u & \text{if } u < 0 \end{cases} \label{eq:pinball2}
\end{align}
%\end{equation}
%TODO explain this
Since QR is directly modelling the quantiles of the target distribution, it does not make assumptions about the underlying distribution of the target. It is thus considered a non-parametric approach. 

\subsubsection{Maximum Likelihood Estimation}\label{sec:MLE}
MLE is a parametric approach, as a underlying assumption about data distribution is being made and used as prior information to be updated based on the learned distribution of the data. 
% Talk about what the output here will be
In practice, the output of the MLE approach are parameters used to create the assumed distribution. Given a standard Gaussian distribution, the output of the model would be the mean and standard deviation parameters. The parameters can then be used to compute a probability distribution for each time-point in the output. Output shape here depends on how many parameters are used to model the target distribution: i.e.~for a Gaussian $y \in \mathbb{R}^{B \times P \times 2}$.
% Talk about the loss function here
MLE is optimized by minimizing the negative log-likelihood of the target distribution shown in Equation~\ref{eq:nll}. Here, $f(y_{b,w}|\hat{y}_{b,w})$ is the likelihood of observing the data $y_{b,w}$ given the model estimations $\hat{y}_{b,w}$.
\begin{equation}\label{eq:nll}
    L_{MLE}(y, \hat{y}) = -\frac{1}{BW}\sum_{b \in B} \sum_{w \in W} \log(f(y_{b,w}|\hat{y}_{b,w}))
\end{equation}

% last sentence again about distirbutions
\subsubsection{MLE Distributions}\label{sec:distributions}
Four different probability distributions are being tested in the MLE approach: Gaussian, Johnson's SU, Johnson's SB and Weibull. 
The target distribution of solar irradiance data is heavily skewed no matter if clear sky values are deducted or not. Hence, the hypothesis is that a distribution as prior in the MLE approach would benefit from being able to fit the skewness of the solar irradiance distribution. In order to test this hypothesis, four different distributions are implemented and evaluated. 
The Gaussian distribution is commonly used in MLE, but not able to model skewness thus utilized as a comparison here. Johnson's SU and SB distribution, and Weibull distribution are both able to model skewness. Weibull has been used in the context of solar irradiance forecasts before, but to our knowledge, Johnson's SU has only been used with wind forecasts~\cite{bentsenRelativeEvaluationProbabilistic2024} while Johnson's SB neither been used with solar or wind forecasts.

The Normal, or Gaussian distribution features two parameters, mean and standard deviation $\mathcal{N}(\mu,\sigma^2)$ with a probability density function (PDF) as seen in Equation~\ref{eq:gaussian}.
\begin{equation}\label{eq:gaussian}
    \mathcal{N}(\mu,\sigma) \sim n(x) =  \frac{1}{\sqrt{2\pi\sigma^2}}e^{-\frac{(x-\mu)^2}{2\sigma^2}}
\end{equation}
The Johnson's family of distributions are transformations on the Gaussian distribution, featuring four parameters $J(\xi,\lambda,\gamma,\delta)$ with which it is able to model any skew and kurtosis, e.g.~such as skew in either negative or positive direction (see~\cite{Edward2004Uniqueness} for a detailed introduction to Johnson's distribution system). Johnson's SU transformation on a Gaussian distribution $n(x)$ as seen in Equation~\ref{eq:johnsonsu} is unbounded, while Johnson's SB transformation (Equation~\ref{eq:johnsonsb}) is bounded between parameter $\xi$ and $\xi + \lambda$.
\begin{align}
    J_{SU}(x) &= \xi + \lambda\sinh\!\left(\frac{n(x)-\gamma}{\delta}\right) \label{eq:johnsonsu} \\
    J_{SB}(x) &= \xi + \frac{\lambda} {e^{-\frac{n(x)-\gamma}{\delta}}+1} \label{eq:johnsonsb}
\end{align}
Since min-max scaling is used on the data, it makes sense to keep the Johnson's SB distribution bounded between zero and one. This means $\xi$ and $\lambda$ are kept fixed at zero and one respectively while only $\gamma$ and $\delta$ are optimized. The injected $cs$ from Equation~\ref{eq:clear_sky_inject} is added to the parameter $\xi$ in Johnson's SU as this closely resembles a mean for Johnson's SU. As $\xi$ is being kept constant in Johnson's SB and representing the minimum of the distribution, $cs$ is instead injected into $\gamma$, which as skew parameter is able to closely influence location of the peak of the distribution.  
Noteworthy here  is that for Johnson's SU, SB and Weibull, the parameters to be optimized are restricted into defined ranges depending on their impact on the distribution. This restriction allows the injected $cs$ to be scaled to the parameter in a meaningful way. Specifically for Johnson's SB, $\gamma$ is restricted to be optimized in the interval $(-4, 4)$, while $\delta$ is restricted to be optimized in the interval $(-2,6)$. The scaled $cs = (1-cs)*4$ is then added to $\gamma$. In Johnson's SU $\gamma$ is restricted to be optimized in the interval $(-4,4)$, while $\delta$ is restricted to be optimized in the interval $(5,9)$.

Weibull features two parameters, shape and scale, $W(\phi,\omega)$ which are able to model the skewness of the data in one direction. The PDF of the Weibull distribution is shown in Equation~\ref{eq:weibull}.
\begin{equation}\label{eq:weibull}
    W(\phi,\omega) \sim w(x) =
    \begin{cases}
    \frac{\omega}{\phi}\left(\frac{x}{\phi}\right)^{\omega-1}e^{-(x/\phi)^\omega} & x \geq 0\\
    0 & x < 0
    \end{cases}
\end{equation}
        
Weibull is bounded, meaning that input and output is usually in the range $y = (0,+ \inf)$, depending on the range of target solutions, this may need some workaround to fit output to the target range. Similarly, $\phi$ and $\omega$ are also bound between $0$ and $+\inf$. See~\cite{kamComparativeWeibullDistribution2021} for a detailed introduction to Weibull distributions.
The bounding issue regarding input data is solved in section~\ref{sec:preprocessing} by using the min-max scaler. The output range however is another issue here as data is normalized before training and only inversely transformed for evaluation. The min-max scaler is extended by adding small values to the final scaled output such that transformed values can not result in a zero. 
During optimization of Weibull, $\phi$ and $\omega$ and restricted to be optimized in the intervals $(0,1)$ and $(0,2)$ respectively. The injected $cs$ is added to $\omega$ as this parameter seems closely related to the mean of the distribution.

\subsection{Baseline Comparisons}\label{sec:baseline}

To justify the inclusion of the high amount of features, a Multi-layer Perceptron (MLP) is trained on only the solar irradiance observations from station eleven. %It is found to perform significantly worse than LSTMs trained on all data features, confirming that useful information are gained from the additional data.

Deterministic results can also be compared to a smart persistence model, such as described in~\cite{klyveValueForecastsPV2023a}. 
\begin{equation}\label{eq:smart_persistence}
   f(t) = \frac{x(t)}{cs(t)}*cs(t+h)
\end{equation}
 Equation \ref{eq:smart_persistence} shows that smart persistence generates projected values based on past values at the same time of day and future clear sky values. Here, $x(t)$ is an observed value at time $t$, $cs(t)$ is a clear sky value at time $t$, and $h$ being a shifting factor.
 Smart persistence can be adapted to a 36 hours horizon by using each of last 24 hours of observations and scale them to the then expected highest irradiance in the next 36 hours (the first half (0-12 hours) of the data are thus used twice from 0-12 hours into the future and 24-36 hours into the future).

\section{Results and Analysis}\label{sec:results}

\subsection{Hyper-parameter Optimization}
 Extrapolating from the results of the deterministic LSTM, the overall hyper-parameters are optimized and applied to all proposed LSTM variations (QR-LSTM and MLE-LSTMs). The whole set of hyper-parameters included input window size, hidden unit size, learning rate, batch size and number of layers. The final LSTM uses a window of three days of past data, two layers, a learning rate of $10^{-5}$, a batch size of 64, and 128 hidden units. 
 
Noteworthy is that optimization was done only on the validation data set. The test data was only used to generate the results presented in the following sections.
\subsection{Model Evaluation}

\begin{table}
    \centering
    \caption{Optimized model results on the test split. MLE-G stand for MLE assuming a Gaussian distribution, MLE-W is using the Weibull distribution, while MLE-JSu and MLE-JSb are using Johnson's SU and SB distributions respectively. MAE, RMSE, and Quantile loss are given in $W/m^2$. Best performers are bold faced.}\label{tab:results}
    \begin{tabular}{ccccc}
         & MAE & RMSE & Quantile-l & ACE \\
       Smart persistence &58.376 &92.539 & - & -  \\
       1-Station-MLP &69.119 & 93.156 & - & -  \\
       % QR-MLP  &  & &  &  &[] \\
       % MLE-G-MLP  &  & &  &  &[] \\
       % MLE-J-MLP  &  & & &   &[] \\
       LSTM  &52.886 & \textbf{77.515}   & -  &-\\
       QR-LSTM  & 53.102 & 80.481 & 211.999&\textbf{0.065} \\
       MLE-G-LSTM & 52.832& 79.904 &208.424 &0.146 \\
       MLE-JSu-LSTM  & 52.499 & 79.878 & 208.472 & 0.121\\
       MLE-JSb-LSTM  & \textbf{48.984} & 79.294 & \textbf{198.324} & 0.309\\
       MLE-W-LSTM  & 56.012 & 88.173 & 228.997 &0.336\\
    \end{tabular}
    
\end{table}

\begin{figure}[t]
    \centering
    \includegraphics[width=0.32\textwidth]{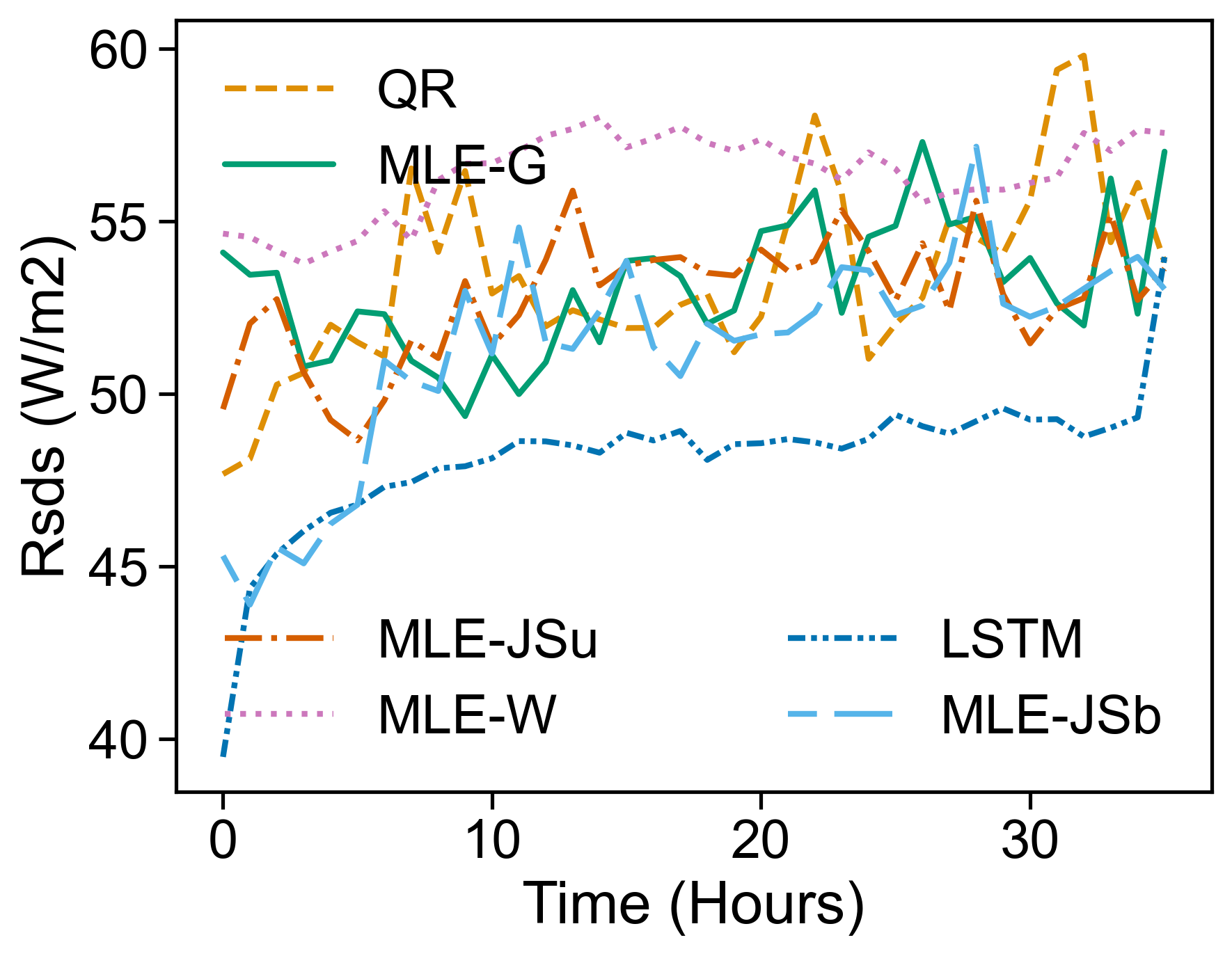}
    \caption{Averaged error distribution of RMSE across the 36 hour output window.}
    \label{fig:error_distributions}
\end{figure}
\begin{figure}[t]
    \centering
    \includegraphics[width=0.32\textwidth]{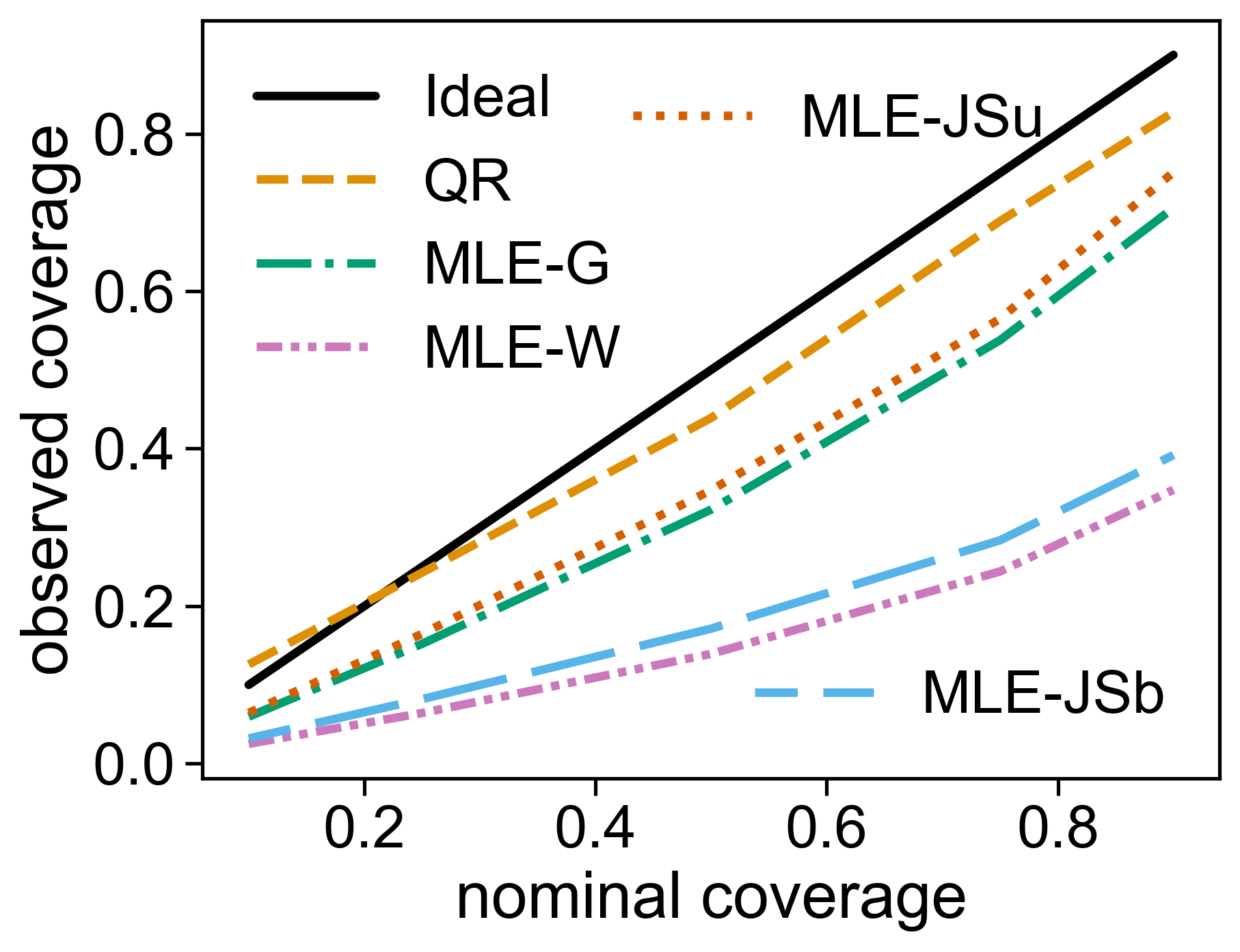}
    \caption{PICP values showing the observed coverage (x) of probability distributions within set intervals (y). Tighter fit to the black line is optimal.}
    \label{fig:picp}
\end{figure}

Testing results of the models in overall test error can be examined in Table~\ref{tab:results}. Performance metrics used here are mean absolute error (MAE), and root mean squared error (RMSE)(see Equation~\ref{eq:rmse_mae}). Additionally, probabilistic approaches feature quantile loss, also called Pinball loss (shown in section~\ref{sec:QR}), average coverage error (ACE), and prediction interval coverage percentage (PICP) shown in Figure~\ref{fig:picp}. 
%(for details on each metric see \cite{bentsenRelativeEvaluationProbabilistic2024}).

\begin{equation}\label{eq:rmse_mae}
    RMSE = \sqrt{\frac{1}{B} \sum_{b=1}^{B} (y_b - \hat{y}_b)^2}, MAE = \frac{1}{B} \sum_{b=1}^{N} |y_b - \hat{y}_b|    
\end{equation}
PICP is one measure of how well prediction intervals of probabilistic models are calibrated. It shows the percentage of true values that fall within the predicted interval. In Equation~\ref{eq:picp}, $s$ represents the count of whether a true value $y$ falls within the predicted interval between the points $[\hat{I}^{upper}_{b,c}, \hat{I}^{lower}_{b,c}]$ for each batch $b$ and confidence interval $c$. As percentage, a good PICP will be as close each $c$ as possible.

\begin{equation}\label{eq:picp}
    PICP_c = \frac{1}{B} \sum_{b=1}^{B}s_b,    s_b = \begin{cases} 1 & \text{if } y_b \in [\hat{I}^{upper}_{b,c}, \hat{I}^{lower}_{b,c}] \\ 0 & \text{otherwise} \end{cases}
\end{equation}
ACE then shows the average difference between the most ideal intervals and the predicted interval coverages. A lower average distance would be ideal here, thus lower ACE are preferred. In Equation~\ref{eq:ACE} $\mathbb{C}$ is the set of ideal confidence intervals.
\begin{equation}\label{eq:ACE}
    ACE = \frac{1}{|\mathbb{C}|} \sum_{c=1}^{\mathbb{C}} |c - PICP_c|
\end{equation}

%Models are trained five times with the same hyper-parameters, but different random seeds, then tested on the test split, in order to show model performance variation and stability with different random initialization of weights and data-presentation order. 

  \begin{figure}[t]
    \centering
    \includegraphics[width=0.32\textwidth]{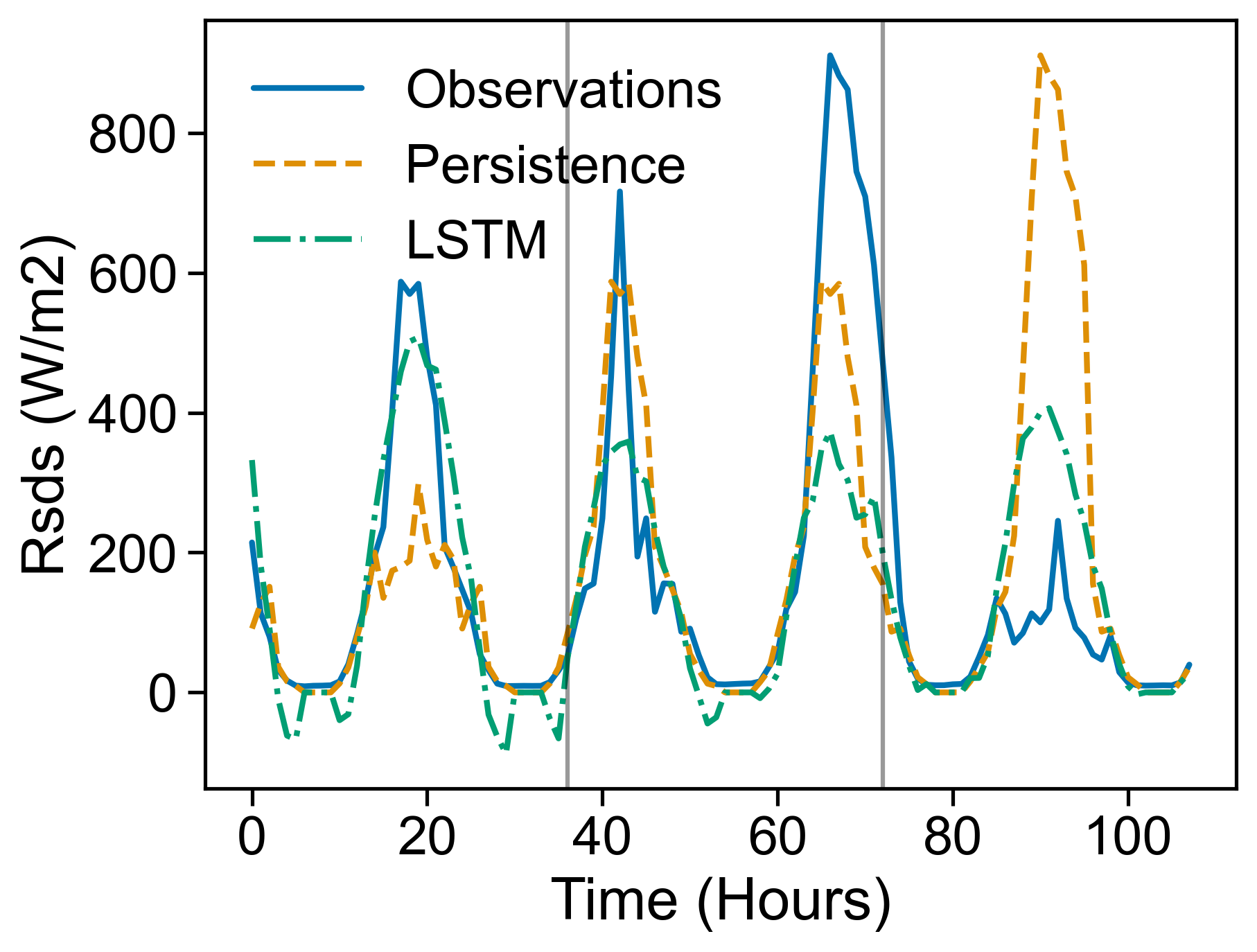}
    \caption{Point prediction results plotted for every 36 hours of forecast. Grey vertical lines designate the beginning of each 36 hours window.}
    \label{fig:results_det}
\end{figure}
\begin{figure}[t]
    \centering
    \includegraphics[width=0.32\textwidth]{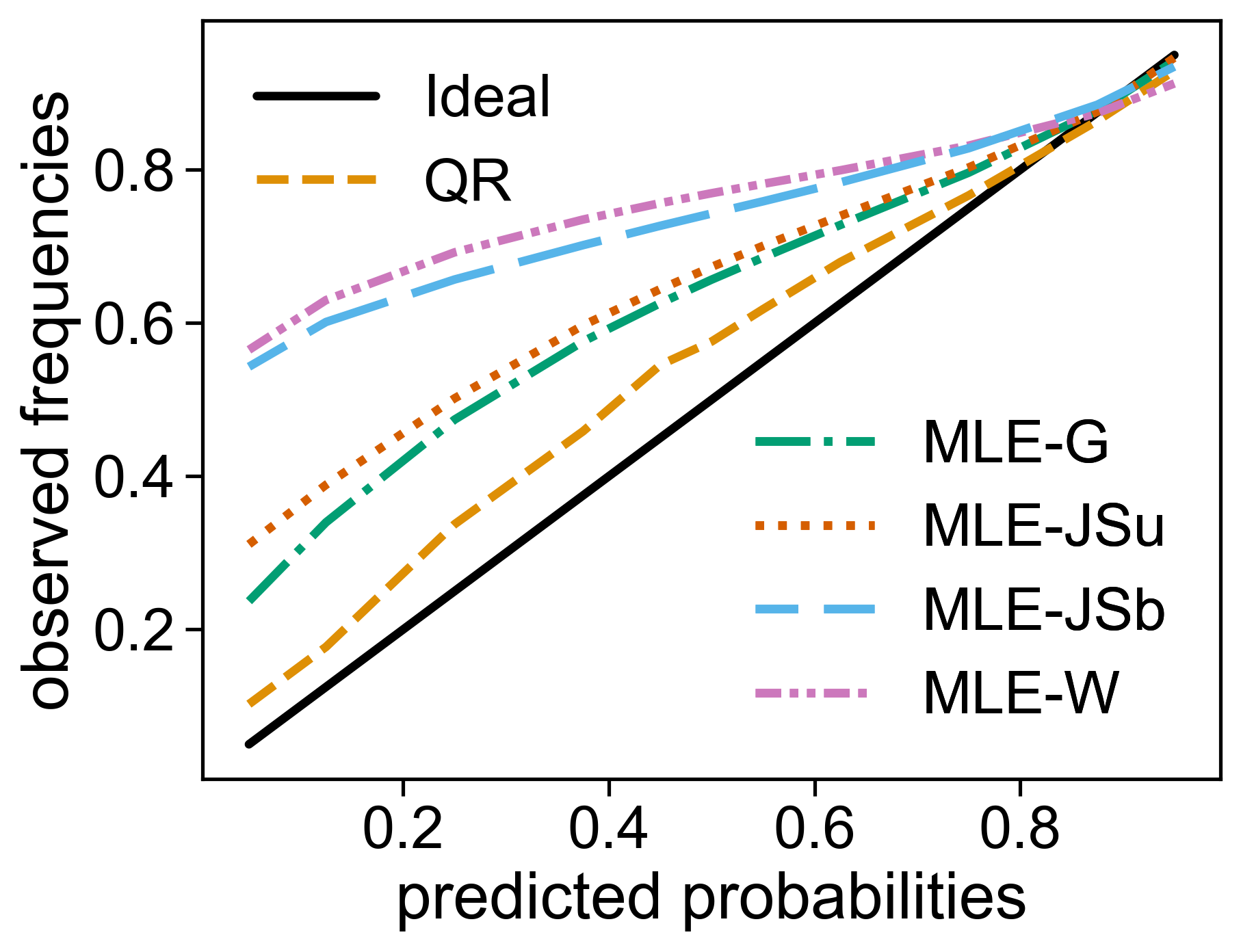}
    \caption{Reliability diagram showing frequency (x) of true observations falling within quantiles (y) of the predicted probability distributions. Tighter fit to the black line is optimal.}
    \label{fig:reliability}
\end{figure}

 The MAE and RMSE of smart persistence are notably worse than all the fitted model performances of Table~\ref{tab:results}. MAE differs not as much, although RMSE punishes higher error more harshly: It is likely that smart persistence fits perfectly during the night, in which LSTMs are often generating small errors, while fitting less well during the day where higher errors are more prevalent. This is probably also the reason why MLE with Johnson's SU, SB, and the Gaussian distribution shows a better MAE than the deterministic LSTM, even though the deterministic LSTM's RMSE is the best out of all tested models. The single station MLP is found to perform significantly worse than LSTMs trained on all data features (apart from Weibull), confirming that useful information are gained from the additional data.
 
 Noteworthy, Figure~\ref{fig:error_distributions} shows the averaged distribution of RMSE of all models across the 36 hour output window: The deterministic LSTM is much less variable here (likely because it is trained on MSE loss), but also shows a large increase in performance in the first two hours and a large decrease in the last one hour. This behavior is is otherwise only visible for Johnson's SB distribution which seems to follow the deterministic LSTM in the first 5 hours. Other probabilistic model only show a weak decrease in performance across the whole 36 hour time window. 

 \begin{figure*}[h!]
    \centering
    \includegraphics[width=0.7\textwidth]{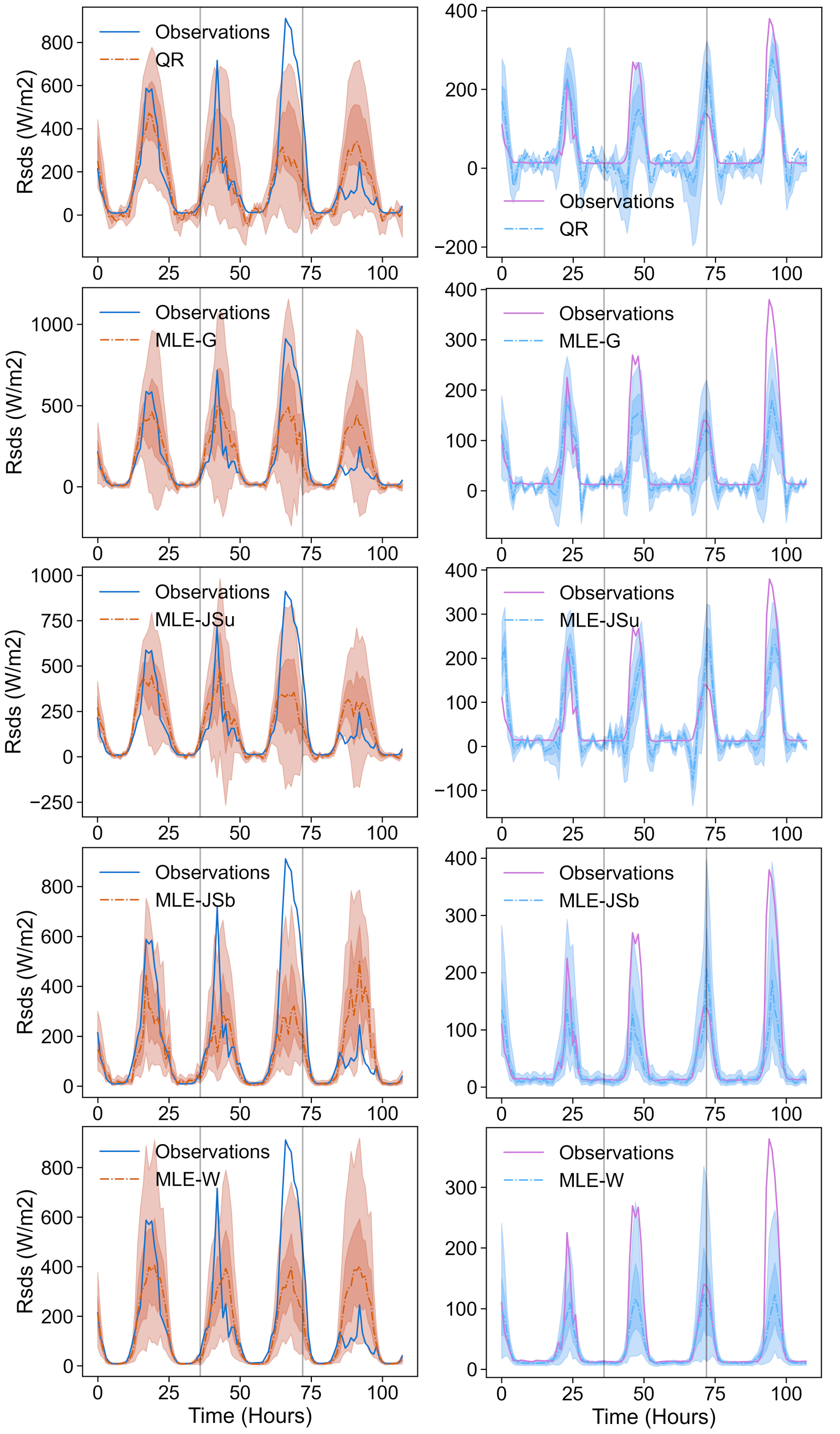}   %{unified_prob2.png}
       \caption{Probabilistic results plotted for every 36 hours of forecast. An example for summer is on the left side, while winter is shown on the right. Output windows of 36 hours are shown by the grey vertical lines. The error bands show the estimated 90\% and 50\% confidence intervals. MLE-G stand for MLE assuming a Gaussian distribution, while MLE-JSu/Jsb stands for MLE, with Johnson's SU and SB distribution respectively.}
       \label{fig:results_prob}
\end{figure*}
Figure~\ref{fig:results_det} shows performance of smart persistence compared to the deterministic LSTM. Smart persistence often over- or underestimates irradiance based on the previous day(s). In comparison, the LSTM does not deviate very much from its estimated average, likely based on clear sky value injection.

 In terms of the probabilistic LSTM performance, MLE-JSb seems to perform best for MAE, RMSE, and quantile loss, but almost worst in ACE. MLE-G seems to perform very similar to the MLE-JSu, with slightly worse prediction errors and a slightly better quantile loss. Only in ACE has the Johnson's SU distribution a clear edge over the Gaussian. 

 Weibull seems to generally do much worse than the other distributions, however this may also be attributed to the training procedure: Injecting clear sky values into parameter outputs of the neural networks makes sense with functions which have a parameter being more closely related to a mean value (such as Gaussian and Johnson's SU), but not with Weibull. 
 
 QR results show a slightly worse performance in MAE, RMSE, and quantile loss, but a significantly better ACE compared to MLE with Gaussian, or Johnson's Sb and SB distribution. It seems that QR is able to show better calibration than MLE approaches (due to its non-parametric nature) with the drawback of slightly worse performance otherwise.
 
 This is reflected in the PICP plot shown in Figure~\ref{fig:picp} and the reliability diagram in Figure~\ref{fig:reliability}. The reliability diagram shows the frequency of true observations being captured within the quantiles of the predicted probability distributions (also see~\cite{lauretVerificationSolarIrradiance2019}). In essence then, in PICP we are examining intervals, while in the reliability diagram we are examining quantiles. In both figures a perfect model would show a straight diagonal line from zero to one. 
 Generally Figure~\ref{fig:reliability} seems to reveal QR as the best calibrated model with unbounded and bounded MLE probability distributions forming groupings of decreasing performance. These groupings are visible in both figures and show a trend towards overestimation of uncertainty in the tails of the distributions. 
 
 It has to be noted that MLE training optimization (especially for Johnson's SU, SB and Weibull) shows a trade-off between ACE on one side and Pinball or RMSE on the other side. 
 Models were generally optimized by their lowest validation loss, however, this was not always the best choice for Johnson's SU, SB, or Gaussian, as at some point, ACE would start to rise while validation loss, quantile loss and point-value errors still continued to shrink. For these cases, an early training stop is performed when ACE is observed to jump up.
In Weibull, all metrics would show a decrease in performance after a while, but minimal Pinball seemed to result in the best overall performance.

 A more comprehensive comparison of the probabilistic models can be seen in Figure~\ref{fig:results_prob}. The figure shows 3 forecasts of 36 hour windows shifted by the window size to achieve a connected plot. The left side shows a test response to summer data (red), while the right side shows the same for winter data (blue).
 All models show similar issues of over- or underestimating low or large irradiance values respectively. It seems like fitting winter is a bit easier for all models and uncertainty seems much lower. Chaotic nightly forecasts are much more visible in winter due to lower maximal values. Unsurprisingly, MLE-W and MLE-JSb are the only one which does not return negative estimations. QR shows more errors during the night compared to MLE models. It also seems like MLE models did a better job at at least including very high values in their confidence intervals compared to QR. However, to reach a full conclusions based on these observations, one would need more to examine than the given plots.     

\section{Discussion}
    % Discussion of the significance of the work for the field
    % Summary of the work
 
Performance of the optimized deterministic LSTM and its probabilistic counter parts trained on solar irradiance observations adjacent to the Arctic circle are better than the comparative baseline models of the single-station MLP and smart persistence. Still, their performance leaves potential for improvement: LSTM results seem too adjacent to injected clear sky values and seasonal differences in performance seems to vary across models. Both may indicate needed changes to model architecture, training process, or preprocessing of data.

While MLE, especially Johnson's SB distribution, seems promising apart from distribution calibration, its training process is not as straightforward as QR. MLE parameter outputs needed to be manually restricted to a reasonable range to achieve increased performance. Based on this, it seems that MLE easily finds and remains in local minima, hence has trouble finding effective distribution parameter values when left alone. %%% TODO COULD ADD RECALIBRATION METHODS
Still, QR does not show a edge in most performance metrics, showing MLE as a viable alternative for probabilistic forecasting. Interesting is that a completely bounded distribution like Johnson's SB seems to perform better than its unbounded alternatives in all metrics, but metrics related to distribution calibration. Intuitively, a restriction in bound that absolutely reflects the target range of the data should be beneficial (here the reminder that the solar irradiance data is min-max scaled to be between zero and one). In terms of calibration, it seems to be that hard bounds are not beneficial anymore, as tails will always extend to zero and one. Following this train of thought, maybe one should allow the optimization of the minimum and maximum parameters, but this would require a more sophisticated optimization process, as normal log-likelihood will fail as soon as true observations exceed the set bounds of the estimated distribution. 

In comparison to other MLE distributions, Weibull seems to be left behind in performance, but this is likely due to the clear sky injection not being as impactful on the distribution parameters as it is on the other distributions.

In the future, the architecture could be specialized (e.g.~using graph neural networks) to encode the spatial information contained within the 20 stations of input. Further, more extensive hyper-parameter optimization may significantly increase performance and clarify differences in models. Maybe a split in training summer and winter data with separate parameter sets, or injecting future NWP model forecasts may increase performance. Addressing the MLE training issues by employing more sophisticated probabilistic optimization techniques, such as Variational Inference may be one way to improve on current results. 

Increasing the apparent difficulty in forecasting solar irradiance further is the chosen 36 hours horizon. While one hour horizons are popular and arguably very easy for machine learning techniques, 24-36 hours is a more practically important interval to forecast which is still in need of improved accuracy (due to its benefit to the day-ahead market). One big challenge here is that standard loss functions will attribute equal loss to all hours in the horizon, leading to a lack of difference in performance of the first hour compared to the last. Developing methods to deal with this may be another avenue for future research.
\section{Conclusion}
This work presented a comparison of deterministic and probabilistic LSTM-based forecasting models of Norwegian solar irradiance data across a window of 36 hours. Employed models were able to beat baseline comparisons. The deterministic LSTM showed the best point-prediction performance with RMSE, while a QR based LSTM showed a superior calibration of uncertainty. However, an MLE models based Johnson's SB distributions showed a great overall performance, with very close point-prediction performance to the regular LSTM. As such, this work shows that MLE can be a viable alternative to QR for probabilistic forecasting. In addition, injecting clear sky values into the output of the LSTMs helped with the seasonal differences in solar irradiance magnitude that otherwise may be difficult to capture with a LSTM.   
\bibliography{refs.bib}
\end{document}